\documentclass[11pt,a4paper]{article}
\usepackage[hyperref]{acl2021}
\usepackage{times}
\usepackage{latexsym}

\usepackage{multirow}

\usepackage{microtype}

\usepackage{graphicx}

\usepackage{hyperref}

\usepackage{amsmath,amsthm,verbatim,amssymb,amsfonts,amscd,mathrsfs}

\usepackage{multirow}
\usepackage{graphicx}
\usepackage[normalem]{ulem}
\useunder{\uline}{\ul}{}

\aclfinalcopy 

\title{Learning ULMFiT and Self-Distillation with Calibration for Medical Dialogue System}

\author{Shuang Ao \\
    Doti Health Ltd \\
  \texttt{ao.shuang@u.nus.edu} \\\And
  Xeno Acharya  \\
  Doti Health Ltd \\
  \texttt{xeno.acharya@gmail.com} \\}

\date{}

\begin{document}
\maketitle
\begin{abstract}
A medical dialogue system is essential for healthcare service as providing primary clinical advice and diagnoses. It has been gradually adopted and practiced in medical organizations in the form of a conversational bot, largely due to the advancement of NLP. In recent years, the introduction of state-of-the-art deep learning models and transfer learning techniques like Universal Language Model Fine Tuning (ULMFiT) and Knowledge Distillation (KD) largely contributes to the performance of NLP tasks. However, some deep neural networks are poorly calibrated and wrongly estimate the uncertainty. Hence the model is not trustworthy, especially in sensitive medical decision-making systems and safety tasks. In this paper, we investigate the well-calibrated model for ULMFiT and self-distillation (SD) in a medical dialogue system. The calibrated ULMFiT (CULMFiT) is obtained by incorporating label smoothing (LS), a commonly used regularization technique to achieve a well-calibrated model. Moreover, we apply the technique to recalibrate the confidence score called temperature scaling (TS) with KD to observe its correlation with network calibration. To further understand the relation between SD and calibration, we use both fixed and optimal temperatures to fine-tune the whole model. All experiments are conducted on the consultation backpain dataset collected by experts then further validated using a large publicly medial dialogue corpus. We empirically show that our proposed methodologies outperform conventional methods in terms of accuracy and robustness.

\end{abstract}

\section{Introduction}
The medical dialogue system is becoming a necessary tool for the doctor-patient interaction as it provides the primary clinical advice and long-distance diagnoses, shortening the checking duration and reducing the manpower cost. It is gradually applied and accepted especially during the pandemic time.

In order to provide an integrated conversational system for back pain management, the system needs to be equipped with evidence on the aforementioned determinants of health. This could best be facilitated by incorporating this evidence through a medical dialogue system.
However, the insufficient medical corpus is one of the biggest restrictions for the training of the neural conversational model. We build the dataset particularly for back pain consultation, including the query and suggestions regarding possible causes, symptoms, and treatment of back pain, which to our best knowledge, is the first medical conversational dataset subjected in the backpain field. 

To achieve a promising accuracy of the sentence generation model, we choose the state-of-the-art transformer model~\cite{vaswani2017attention} as the benchmark. As transfer learning has shown great success in machine learning tasks such as classification and regression~\cite{pan2009survey}, in this paper, we choose two well-known and efficient techniques: Universal Language Model Fine Tuning (ULMFiT)~\cite{howard2018universal} and Knowledge Distillation (KD)~\cite{hinton2015distilling}. ULMFiT provides additional help to improve the model accuracy by transferring information from language modeling to NLP downstream tasks, such as conversational model, sentiment analysis, and Machine Translation. Due to its obvious advantages, we implement the pretrained language model on top of the conversation model to get better feature extraction. Furthermore, KD transfers the knowledge from a cumbersome model to a lighter-weight model so that the small model can replicate the result. KD has been used in the recent NLP research, such as text classification and sequence labeling~\cite{yang2020textbrewer} and got the promising result. However, due to the limitation of data size and robust model, the application KD is not flexible to some extent. To resolve these issues, Self Distillation (SD) ~\cite{yuan2019revisit} is proposed, where the student model is used as the teacher model as well. Results show that SD can almost replicate the accuracy regardless of a well-trained large model or big dataset such as in the image classification task~\cite{zhang2019your}. SD has also been applied to NLP tasks such as language model and neural machine translation~\cite{hahn2019self} and obtains promising results.

Despite obtaining higher accuracy and better performance, modern deep learning models face drawbacks of miscalibration and overconfidence ~\cite{muller2019does, naeini2015obtaining, lakshminarayanan2016simple}. Recent studies resolve this issue by using techniques like label smoothing~\cite{muller2019does} and temperature scaling~\cite{naeini2015obtaining}, and Dirichlet calibration~\cite{kull2019beyond}. These works show that the well-calibrated model can improve the model performance as well as feature representation. As for the NLP downstream tasks, research has shown that calibration benefits both sentence quality and length in the sentence classification~\cite{jung2020posterior}, and helps to improve the model fine-tuning in text generation~\cite{kong2020calibrated}. 

As transfer learning techniques and calibration contributes to NLP tasks, we investigate the correlation of improving calibrated feature representation with ULMFiT and SD. Label smoothing is integrated with ULMFiT to extract significant features from language modeling. To improve KD by recalibrating predicted probability, we incorporate temperature scaling (TS) with knowledge distillation loss. We also observe the correlation of a well-calibrated trained network in whole model fine-tuning. We conduct extensive experiments to validate our observations with two datasets of (1) the consultation back pain and (2) medical dialogue. Results show that a well-calibrated model is highly correlated with ULMFiT and SD, as well as fine-tuning, in terms of both accuracy and calibration error. 

Our contributions can be concluded as following:\\
(1) We introduce the calibrated ULMFiT (CULMFiT) by applying label smoothing on conventional ULMFiT. Results are showing that the CULMFiT outperforms the vanilla ULMFiT, proving the impact of calibration of language modeling. \\
(2) We measure optimal calibrated temperature and replace the fixed temperature value in KD loss and demonstrate that calibrated temperature outperforms the fixed value.\\
(3) We incorporate temperature scaling with the whole model fine-tuning and observe that calibration benefits model performance and uncertainty. \\
(4) We build the consultation backpain dataset, consisting of patients' queries and clinicians' responses into conversational pairs.



\section{Proposed Method}
\subsection{Preliminaries}
\paragraph{ULMFiT}
Natural Language Processing has picked up the pace in recent years and caught researchers' attention greatly, essentially attributed to the conquer of inductive transfer learning, which was seen as the major obstacle that NLP was lagging behind Computer Vision (CV). Universal Language Fine Tuning (ULMFiT) was proposed~\cite{howard2018universal} as obtaining the success of passing the acquired knowledge of pre-trained model to other similar tasks.
ULMFiT is to pretrain the model on a large general domain corpus such as Wikipedia data, then fine-tune it on the target tasks. As a source task trained with a large corpus, the pre-trained language model can capture most facets and contexts of the data, which is ideal for NLP downstream tasks. Hence including Text Classification that ULFMiT was firstly introduced with, it gets great success and applied in almost all NLP fields. It is believed that with the language model trained on the large-scale data, the model with small or medium data will also replicate similar results to the vanilla model. 
\paragraph{Label Smoothing (LS)}
Label smoothing has been widely applied in various fields of deep learning, such as image classification~\cite{real2019regularized} and speech recognition~\cite{chorowski2016towards}. It achieves promising results since Szegedy et al.~\cite{szegedy2016rethinking} first introduced it, then gets further development after the extension explanation on its mechanism of how it improves the model calibration~\cite{muller2019does}. As the regularization technique to tackle the overconfidence of a model, label smoothing softens the one-hot labels in the penultimate layer’s logit vectors, to improve the calibration and further help the robustness and reliability of the model. Here is the mathematical illustration of label smoothing: suppose $\hat{p}_c$ is the probability and $p_c$ is the ground truth of the $c$-th class, where $p_c$ is 1 for the correct class and 0 for the rest classes, the cross-entropy loss of network trained with a hard target can be demonstrated as: $CE = -\sum_{c=1}^{C}p_c\log(\hat{p}_c)$.For a network trained with a label smoothing hyper-parameter $\alpha$, the one-hot true value will be clipped as: $p^{LS}_c = p_c(1-\alpha) + \alpha/C$.
Hence the cross-entropy loss with label smoothing can be illustrated as:
\begin{equation}
CE^{LS} = -\sum_{c=1}^{C}p^{LS}_c\log(\hat{p}_c).
\end{equation}

\paragraph{Temperature Scaling (TS)}
It has been observed that most of the modern neural networks are poorly calibrated even with a high confidence score. To solve this issue and make the model better calibrated, among all possible factors that may influence the calibration, temperature scaling (TS), as a straightforward extension of Platt Scaling, has been verified as the most efficient and least time-consuming and computationally expensive way~\cite{guo2017calibration}. A single scalar T (T$>$1) called temperature is applied on the logit then it passes to the softmax function (denoted as $\sigma$), which will not change the maximum value in it, so the prediction remains intact. Here is the equation for TS given the logit vector:
\begin{equation}
\label{temp}
\hat{p_c}^{TS} = \max_c \: \sigma_ ( logit_c / T)^{(c)}.
\end{equation}

\paragraph{Self-Distillation (SD)}
Knowledge distillation (KD) targets compressing a cumbersome teacher model into a lighter-weight student model. The distilled model can still replicate similar or better accuracy due to the privileged information captured by the teacher model. Suppose the logits for teacher model and student model are $logit^T$ and $logit^S$, and fixed T value as $T^{fix}$, the loss function with of Kullback-Leibler divergence (KL divergence) $L_{KD}$ can be formulated as: 
 
\begin{align}
\label{eq:knowledge_distillation}
L_{KD}=
\sum \mathrm{KL}\Big(\sigma\big(\frac{logit^T}{T^{fix}}\big), \sigma\big(\frac{logit^S}{T^{fix}}\big)\Big) 
\end{align}

It is generally believed that the teacher model should be well-trained with a large corpus and has a bigger capacity than the student model. However, the insufficiency of the dataset and the untrustworthiness of the model are substantial restrictions to KD. Yuan et al~\cite{yuan2019revisit} argue that the student model can achieve similar results with a poor-trained or smaller teacher model, even under the circumstance of no teacher model, which is called self-distillation (SD). By making the model be their own teacher, SD is to train the student model first to get a pre-trained model, then using it as the teacher to train itself. It has been further proved the positive effect that self-distillation has on calibration~\cite{zhang2020self}. 

\subsection{ULMFiT with Label Smoothing}
ULMFiT has obtained great success in NLP tasks as it transfers information from the pre-trained model to the target application domain, and LS helps in calibration and better uncertainty. We apply LS to ULMFiT to gain a calibrated ULMFiT (CULMFiT) to further improve the feature representation and extract more distinctive information from language modeling. Given $\theta^{ULMFiT}$ is the pre-trained ULMFiT weight, x as the input of the conversational model, the loss function of ULMFiT with LS can be written as follows:

\begin{equation}
\label{temp}
CE_U^{LS} = -\sum_{c=1}^{C}p^{LS}_c\log(\hat{p_c}|x,\theta_U). 
\end{equation}

\begin{table*}[]
\centering
\caption{Samples of the backpain dataset. }
\label{table:sample backpain data}
\resizebox{\textwidth}{!}{%
\begin{tabular}{|c|c|l|}
\hline
ID                 &        & \multicolumn{1}{c|}{Medical Dialogue}                                                                                                                                                                                              \\ \hline
\multirow{2}{*}{0} & Enquiry & What is musculoskeletal pain condition?                                                                                                                                                                                            \\ \cline{2-3} 
                   & Reply  & \begin{tabular}[c]{@{}l@{}}A great change of lifestyle and   behaviour, such as too much workload, adjustments in the workplace, \\      work breaks and sudden exercise would improvement of musculoskeletal   pain.\end{tabular} \\ \hline
\multirow{2}{*}{1} & Enquiry & Why my foot pain cause back pain?                                                                                                                                                                                                  \\ \cline{2-3} 
                   & Reply  & The possible reason is your spine’s alignment or overstressing lower back   muscles                                                                                                                                                \\ \hline
\multirow{2}{*}{2} & Enquiry & The back pain cause me unable to carry groceries, what should I do?                                                                                                                                                                 \\ \cline{2-3} 
                   & Reply  & \begin{tabular}[c]{@{}l@{}}Try the grocery delivery or ask   help from your close family or friends. If it is severe,\\      contact your clinician immediately.\end{tabular}                                                      \\ \hline
\multirow{2}{*}{3} & Enquiry & Will back pain influence the enjoyment between couples?                                                                                                                                                                             \\ \cline{2-3} 
                   & Reply  & Yes, studies have shown that higher lever of back pain can impair the   leisure activities with the spouse.                                                                                                                         \\ \hline
\multirow{2}{*}{4} & Enquiry & I feel pain in my joints after exercise, what is the problem?                                                                                                                                                                      \\ \cline{2-3} 
                   & Reply  & \begin{tabular}[c]{@{}l@{}}If your joint feels particularly   painful afterwards for longer than two hours after an exercise session,   \\      reduce the intensity of your next exercise session.\end{tabular}                   \\ \hline
\end{tabular}%
}
\end{table*}

\subsection{Self-Distillation with TS}
Self-distillation (SD) has been proved to replicate the similar accuracy as the knowledge distillation (KD) with the teacher model training on student model, and temperature scaling helps to prevent miscalibration. We integrate TS on SD to attain a well-calibrated distilled model. For this purpose, we adopt KD loss of KL divergence with calibration as in the paper~\cite{hinton2015distilling}. However, temperature set as a scalar value is a similar technique as network calibration, and the optimal temperature is expected to be a better option.  
In our work, we measure optimal T and assign it to the KD, aiming at preventing inappropriate calibration and investigating the relation between calibration and SD. Suppose the logits for the teacher model and student model are $logit^T$ and $logit^S$, and the optimal temperature is $T^{opt}$. The loss function with KL divergence $L_{KD}$ can be formulated as:

\begin{align}
\label{eq:dark_knowledge_distillation}
L_{SD}=
\sum \mathrm{KL}\Big(\sigma\big(\frac{logit^T}{T^{opt}}\big), \sigma\big(\frac{logit^S}{T^{opt}}\big)\Big) 
\end{align}

The final loss $L$ can be demonstrated as:
\begin{equation}
\label{temp}
L = L_{SD} + L_{CE}
\end{equation}

\subsection{Fine-tuning with TS}
As an approach of transfer learning, fine-tuning can propagate the acquired knowledge from one domain to another and enhances the learning capacity. On the other hand, TS produces a well-calibrated confidence score. To further improve the information transformation and feature representation, we apply TS to the logit for cross-entropy loss calculation while fine-tuning the entire model. Given $p_c^{TS}$ is the temperature scaled logit (as shown in formula 2), the loss function with TS can be illustrated as:
\begin{equation}
\label{temp}
CE^{TS} = -\sum_{c=1}^{C}p_c\log(\hat{p}_c^{TS}).
\end{equation}
\section{Experiments}

\begin{table*}[!h]

\centering
\caption{Results of Backpain Dataset. Annotations of experimental models are as following: the vanilla transformer model and ULMFiT are labeled as Baseline; ULMFiT with label smoothing as CULMFiT; model with ULMFiT and fine tune with TS as Fine-tune. }

\label{table:backpain}
\begin{tabular}{|c|c|c|c|c|c|}
 \hline
 Method & Model & BLEU-1 & Perplexity & METEOR & ECE \\   \hline
 \multirow{2}{1.5cm}{Baseline} & Transformer &0.4292  &7.9895 &0.4079&0.3702\\ \cline{2-6}
 & ULMFiT & 0.4321 &8.0603 & 0.4218&0.3764\\   \hline
 LS & CULMFiT & \textbf{0.4632} & 5.6155& \textbf{0.4552}&0.3674\\   \hline
 TS & Fine-tune &0.4415  &\textbf{5.2797} &0.4268 &\textbf{0.2884}\\   \hline
\end{tabular}
\end{table*}


\begin{table*}[!h]
\centering
\caption{Results of MedDialog Dataset: the vanilla transformer model and ULMFiT are annotated as Baseline; ULMFiT is the Transformer model trained with the Medical Dialogue Dataset; regularized ULMFiT is annotated as CULMFiT; the proposed model fine tuned with TS is Fine-tune. }
\label{table:medical dialogue English}
\begin{tabular}{|c|c|c|c|c|c|}
 \hline
 Method & Model & BLEU-1 & Perplexity & METEOR & ECE \\   \hline
 \multirow{2}{1.5cm}{Baseline} &Transformer & 0.3387 & 11.5422 &0.2280 &0.2611\\   \cline{2-6}
 &ULMFiT &0.3609  & \textbf{7.9134}&0.2556&0.3519\\   \hline
 LS &CULMFiT & \textbf{0.3765} & 10.2346& 0.2578&0.3734\\   \hline
 TS &Fine-tune& 0.3747 & 12.6997 &\textbf{0.2618}&\textbf{0.0580}\\   \hline
\end{tabular}
\end{table*}


\subsection{Datasets}
\paragraph{Backpain Dataset}
To develop an evidence-based skillful conversational model, we collect the backpain dataset with pairs of the query from a patient and the response from a clinician. Table~\ref{table:sample backpain data} shows samples of conversational pairs. Sources of queries are various sites people would generally ask health-related questions, such as Google and Quora, and responses are collated from either peer-reviewed journal articles~\cite{hayden2005exercise}~\cite{henschke2010behavioural}~\cite{cagnie2007individual}~\cite{scheermesser2012qualitative}~\cite{choi2010exercises}~\cite{van2018mind} or other sources recognized for providing valid health advice and suggestions like NHS website~\footnote{https://www.nhs.uk/conditions/back-pain/}. It covers five highly related factors that cause back pain, namely sleep, mental health, exercise, nutrition, and social and environmental factors.
The dataset contains 1000 conversational pairs for the train set and 200 pairs for the validation set.

\paragraph{MedDialog}
Due to the disadvantages of the small volume of our backpain dataset, we also use the MedDialog Dataset~\cite{zeng2020meddialog} to further testify our hypothesis of calibration. It consists of conversational pairs of symptoms description from patients and follow-up questions and diagnoses from doctors, which covers various medical fields such as pathology and family medicine. We randomly divide the dataset into train and validation set with the ratio of 0.8 and 0.2. 

\subsection{Implementation Details}

We choose the well-known transformer model as the benchmark in our project. The language modeling architecture for ULMFiT is the encoder part of the Transformer with Fully-Connected (FC) Layers, and the loss function is cross-entropy loss with label smoothing. To fine-tune the proposed model, we first get the optimal TS value, then apply it to recalibrate the logit for the trained model. The GPU of Nvidia Tesla T4 with the memory of 16GB is used to conduct all the experiments in this work. The dataset is split with 0.8 and 0.2 for training and validation. All experiments are conducted with the Adam optimizer, 0.01 as the learning rate and batch size of 4. The best BLEU-1 score metric is used to find the best epoch.

\section{Experiments}

\subsection{Results}

\subsubsection{Evaluation Metrics}
We use the uni-gram similarity metrics BLEU-1 as the major evaluation for our dialogue system. To measure the word overlapping between the ground truth and prediction, we also apply Metric for Evaluation of Translation with Explicit Ordering (METEOR) metric~\cite{banerjee-lavie-2005-meteor} in our work. Perplexity, as the measurement of model uncertainty to the training data, is calculated based on the cross-entropy loss for each sample. We use the Expected Calibration Error (ECE)~\cite{naeini2015obtaining} to check the efficiency of calibration techniques. ECE divides predictions into N equally-spaced bins and takes the weighted mean of each bin's confidence gap. We choose N=15 bins in our work. 


\begin{table*}[!h]
\caption{Visualization of predicted responses. Query is the input and GT is the ground truth. Prediction is the response generated by the model. }
\label{table:visulizationof generated responses}
\centering
\begin{tabular}{|c|c|c|p{6cm}|}
\hline
& Sample &  &Prediction \\   \hline
 \multirow{20}{*}{\rotatebox[origin=c]{90}{Backpain}} & \multirow{10}{6cm}{Query: What to do to solve long time sitting issue except exercise? \\\vspace{\baselineskip}GT: stand up and move about gently for a short period every hour would help relieve the muscle stiffness} &Transformer & up and move for a short time would help the stiffness.\\  \cline{3-4}
 
 &  &ULMFiT & be up and move for a short time every hour would help muscle stiffness.\\   \cline{3-4}
 &  &CULMFiT &stand up and move around gently for a short period every hour would help to solve the muscle stiffness.\\   \cline{3-4}
 &  & Fine-tune&be up and move gently for a short period every hour would help the muscle stiffness.\\   \cline{2-4}
 

 
 \multirow{20}{*}{\rotatebox[origin=c]{90}{MedDialog}} & \multirow{15}{6cm}{Query: Hi doctor, I am 25 years old and I have a history of hair fall for almost 5 years. I am very concerned of it. Could you give me some advice and possible treatment? \\\vspace{\baselineskip}GT: Hi, as per you case history of hair fall, my treatment advice is to take good nutritious diet full of green leafy vegetables and milk, and to use a good herbal shampoo and coconut hair oil.} &Transformer &Hello, per your case of hair fall, my advice be good diet with vegetables and milk, use shampoo and oil out of it. \\   \cline{3-4}
 &  &ULMFiT &Hello, per your case history of hair fall, my advice be take good herbal diet full of green leafy vegetables and milk, use a good shampoo and oil for it. \\   \cline{3-4}
 &  &CULMFiT & Hi, per you case history of hair fall, my treatment advice is to take nutritious diet of green vegetables and milk, and to use a good herbal shampoo and hair oil.\\   \cline{3-4}
 &  & Fine-tune&Hi, per you case history of hair fall, my treatment advice is to take herbal diet of green vegetables and milk, and to use a good herbal shampoo and green herbal oil.\\   \hline

\end{tabular}
\end{table*}

\begin{figure}[!h]
\centerline{\includegraphics[width=0.5\textwidth]{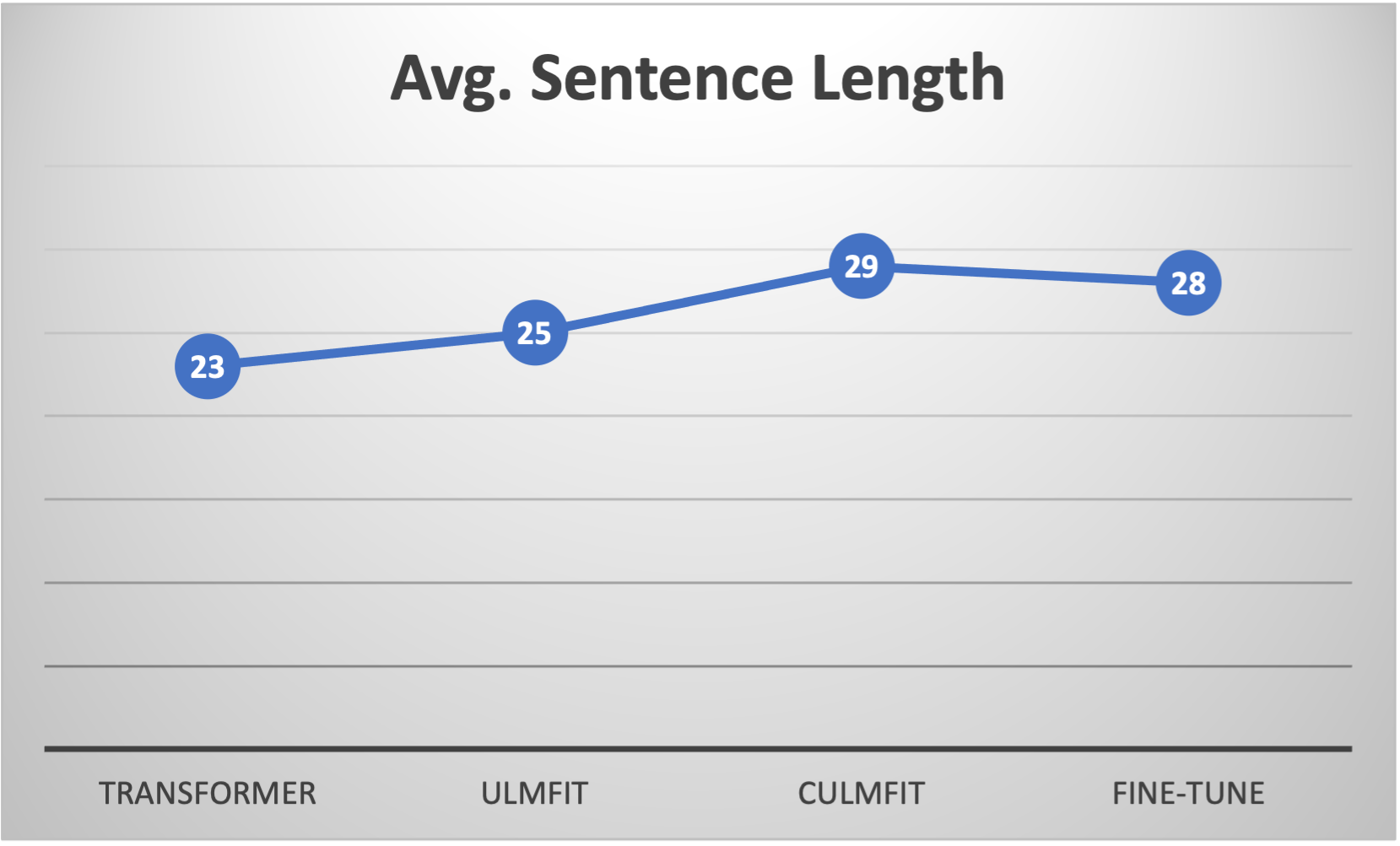}}
\caption{Average sentence length generated by each model in the backpain dataset.}
\label{fig2}
\end{figure}

\subsubsection{Evaluation of the Backpain Dataset}

The results of the dialogue system trained with the consultation backpain dataset are shown in table \ref{table:backpain} and examples of generated responses are demonstrated in table \ref{table:visulizationof generated responses}. The calibrated ULMFiT with LS (CULMFiT) significantly outperforms the baseline transformer model by improving the BLEU-1 score by about 3.8\%, and exceed the vanilla ULMFiT by approximately 1.5\%. On the other hand, the fine-tuning TS improves both BLEU-1 score and ECE with 1\% and 8\%, respectively. Though the fine-tuning with TS does not provide the best BLEU-1 score, it provides the best calibrated confidence score with the lowest ECE. In terms of generated sentence length and quality, our proposed models of CULMFiT and fine-tuning outperform baseline models of transformer and ULMFiT. The diagram \ref{fig2} illustrates that on average, the generated conversation length of CULMFiT and fine-tuning is longer than those from the benchmark models, where CULMFiT model produces the longest responses. Furthermore, the proposed models generate more logical and meaningful sentences. For example, in the first sample in table \ref{table:visulizationof generated responses}, the CULMFiT network predicts the verb "\textit{stand}" and the fine-tuning model generates the phrase "\textit{a short period every hour}" that exactly matches the ground truth, which makes the response more accurate for the symptom.
Overall, CULMFiT demonstrates the best performance on most of the evaluation metrics. Evaluation results prove the effectiveness of proposed calibration techniques with the ULMFiT and fine-tuning on the probability and correctness adjustment.



\begin{table*}[!h]
\centering
\caption{Results of self-distillation with Backpain dataset. Three methods are applied in this experiment: without self-distillation (standalone), self-distillation with a fixed value of TS (SD Fixed TS), and self-distillation with optimal TS (SD optimal TS). The fixed TS is 2. The optimal TS for the transformer model and CULMFiT is 3.025 and 4.789 respectively.}
\label{table:Results of self distillation}
\begin{tabular}{|c|c|c|c|c|c|}
 \hline
 Method & Model & BLEU-1 & Perplexity & METEOR & ECE \\   \hline
 \multirow{2}{1.5cm}{\footnotesize Standalone} & Transformer &0.4292  &7.9895 &0.4079&0.3702\\ \cline{2-6}
 & CULMFiT & \textbf{0.4632} & \textbf{5.6155}&\textbf{0.4552}&0.3674\\   \hline
 \multirow{2}{1.5cm}{\footnotesize SD Fixed TS} & Transformer &0.4331  &7.8329 &0.4221&0.3820\\ \cline{2-6}
 & CULMFiT & 0.4236 & 6.3934&0.4135&0.1962\\   \hline
 \multirow{2}{1.5cm}{\footnotesize SD Optimal TS} & Transformer &0.4334  &7.8010 &0.4187&0.3703\\ \cline{2-6}
 & CULMFiT & 0.4473 & 5.8486 &0.4402&\textbf{0.1788}\\   \hline

\end{tabular}
\end{table*}

\subsubsection{Evaluation of MedDialog Dataset} 
To further verify the hypothesis that calibration benefits the model performance in both accuracy and robustness, we replicate the previous experiments on the Medical Dialogue Dataset. The results of various evaluation metrics are illustrated in table \ref{table:medical dialogue English}, and the sample visualization is shown in \ref{table:visulizationof generated responses}. All results are mostly consistent with the previous experiments. For example, in the sample illustrated in the table \ref{table:visulizationof generated responses}, the length of predicted sentences from CULMFiT and fine-tuning model is longer than the baseline models. Besides, the adjective "\textit{herbal}" for the noun "\textit{shampoo}" from the proposed models can better explain the type of shampoo product, which makes the response more specific for the patient's inquiry. Overall, the proposed methodologies illustrate superior performance in most of the evaluation metrics. The calibrated ULMFiT (CULMFiT) with LS outperforms the benchmark and the vanilla ULMFiT by about 4\% and 1.5\% increment of BLEU-1 score correspondingly. The fine-tuning with the TS model significantly improves ECE by about 35\%. Results from both experiments prove that calibration techniques of LS and TS help to improve the robustness and uncertainty of the model. 

\subsubsection{Evaluation of Self-Distillation With TS} 
\begin{figure}[!h]
\centerline{\includegraphics[width=0.5\textwidth]{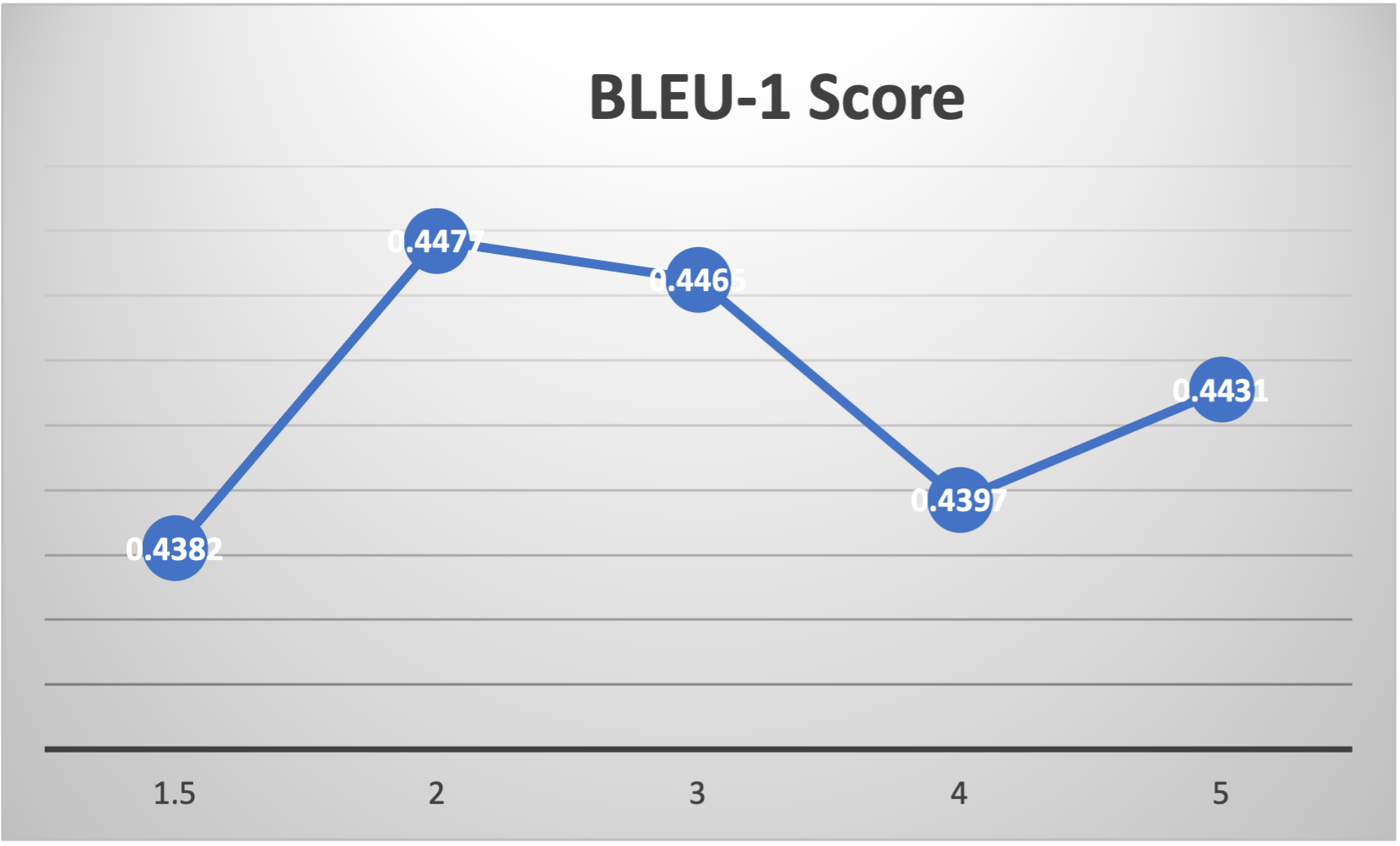}}
\caption{BLEU-1 score with different T values.}
\label{fig1}
\end{figure}

One of our observations is that the SD model with the optimal TS outperforms the one with fixed TS. All results are shown in table \ref{table:Results of self distillation}. We select the benchmark transformer model and the model with the calibrated ULMFiT in this experiment. It has been shown that SD with the optimal T value obtains better performance than with the fixed T (with $T=4$) value for image classification~\cite{hinton2015distilling}. Hence in our work, we also compare the SD with fixed T and optimal T applied in both benchmark and proposed model. To select the best fixed T value, we apply T values of 1.5, 2, 3, 4, and 5 and choose the one with the best BLEU-1 score. The diagram~\ref{fig1} indicates that $T=2$ provides the best BLEU-1 score. Compared to the standalone, SD with fixed and optimal T of transformer and CULMFiT models in table \ref{table:Results of self distillation}, CLUMFiT without SD obtains the best BLEU-1 score, perplexity, and METEOR, while SD with optimal TS provides the best ECE. On the other hand, CULMFiT gets hampered with calibration, which has been evinced in the work~\cite{muller2019does}. Overall, the performance of the model trained with optimal TS beats the one with fixed TS. 

\section{Discussion}
In this paper, we apply calibration techniques of LS and TS to develop the medical dialogue system and get promising results. Table \ref{table:backpain} and \ref{table:medical dialogue English} are showing results that the well-calibrated model benefits ULMFiT, SD and fine-tuning. Table \ref{table:Results of self distillation} demonstrates the observation of self-distillation on fixed and optimal temperature scaling. All our observations is presented with the sample visualization in table \ref{table:visulizationof generated responses}.
Overall, the ULMFiT with LS provides the best BLEU-1 score and the fine-tuning TS improves the ECE mostly, which is consistent with experiments in both datasets. Despite the higher model performance in both accuracy and calibration, fine-tuning is a two-stage training, which can cause an additional computational burden. Even though LS and TS introduce additional computational parameters, the computational cost is negligible. On the other hand, ULMFiT with label smoothing hurts SD, which has been reported in~\cite{muller2019does}.

\section{Conclusion}
In this paper, we propose the calibrated ULMFiT, self-distillation and fine-tuning to build a medical dialogue system. Label smoothing and temperature scaling are utilized to obtain calibrated network and improve the performance in terms of accuracy and robustness. We empirically demonstrate calibration is highly co-related with ULMFiT, SD and fine-tuning, which has been presented in table \ref{table:backpain}, \ref{table:medical dialogue English},\ref{table:visulizationof generated responses} and \ref{table:Results of self distillation}. For future work, we will explore the calibration and knowledge-distillation impact on other NLP downstream tasks like Neural Machine Translation and Sentiment Analysis.




\bibliographystyle{acl_natbib}

\bibliography{references}

\end{document}